\documentclass[%
 reprint,
 amsmath,amssymb,
 aps,
]{revtex4-2}
\usepackage{xcolor}
\usepackage{multirow}
\usepackage{graphicx}
\usepackage{dcolumn}
\usepackage{bm}
\usepackage{subcaption} 
\usepackage{adjustbox}
\usepackage{hyperref}

\begin{document}

\preprint{APS/123-QED}

\title{Granger Causality Detection with Kolmogorov-Arnold Networks}

\author{Hongyu Lin$^1$}
\email{Corresponding author \\
Email address: hongyu.lin.18@ucl.ac.uk}

\author{Mohan Ren$^1$}
\author{Paolo Barucca$^1$}
\author{Tomaso Aste$^{1,2}$}

\affiliation{$^1$Department of Computer Science, University College London, London, United Kingdom.}
\affiliation{$^2$Systemic Risk Centre, London School of Economics, London, United Kingdom.}

\begin{abstract}
Discovering causal relationships in time series data is central in many scientific areas, ranging from economics to climate science. Granger causality is a powerful tool for causality detection. However, its original formulation is limited by its linear form and only recently nonlinear machine-learning generalizations have been introduced. This study contributes to the definition of neural Granger causality models by investigating the application of Kolmogorov-Arnold networks (KANs) in Granger causality detection and comparing their capabilities against multilayer perceptrons (MLP).  In this work, we develop a framework called Granger Causality KAN (GC-KAN) along with a tailored training approach designed specifically for Granger causality detection. We test this framework on both Vector Autoregressive (VAR) models and chaotic Lorenz-96 systems, analysing the ability of KANs to sparsify input features by identifying Granger causal relationships, providing a concise yet accurate model for Granger causality detection. 
Our findings show the potential of KANs to outperform MLPs in discerning interpretable Granger causal relationships, particularly for the ability of identifying sparse Granger causality patterns in high-dimensional settings, and more generally, the potential of AI in causality discovery for the dynamical laws in physical systems. 
\end{abstract}

\maketitle

\section{Introduction}
In time series analysis, Granger causality, first introduced by \cite{granger1969}, has long been a fundamental tool in statistical method for exploring causal relationships between time series. Unlike static correlation analysis, Granger causality examines temporal relationships, assessing whether the historical information of one time series can improve the prediction of another, while accounting for the predictive information contained in the history of the latter. Despite ongoing debates about the general applicability of this framework, it  serves as an operational framework that allows causality to be inferred in a data-driven manner from time series.

Granger causality originally focuses on bivariate analysis under the assumptions of linear dependency and stationary time series. Specifically, if the past values of one time series $x_i$ can help predict another series $x_j$ beyond what is achieved by the past of $x_j$ alone, then $x_i$ is said to Granger-cause $x_j$.  
In the general multivariate setting, Vector Autoregressive (VAR) models have been used, where multiple time series and their lags are jointly modeled to infer Granger-causal relationships \cite{basu2015network, chen2004analyzing, barrett2010multivariate}. Moreover, extensions to nonlinear dynamics, such as the nonlinear Autoregressive (NAR) model \cite{billings2013nonlinear}, overcome the linearity assumption by incorporating a nonlinear function which takes the past values of all the time series as arguments.

To address the complexity of nonlinear and high-dimensional time series, several researchers have utilized neural networks to model Granger causality. Tank et al. \cite{Tank_2021} introduced the component-wise Multilayer Perceptron (cMLP) and the component-wise Long Short-Term Memory network (cLSTM) for multivariate nonlinear Granger causality. In these approaches, each target time series is independently modeled using a dedicated MLP or LSTM, with the lagged values of all time series serving as inputs. Granger causality is then inferred by analyzing the weights connecting the input layer to the first hidden layer, where significant weights indicate Granger causal relationships. Nauta et al. \cite{nauta2019causal} proposed the Temporal Causal detection Framework (TCDF), which employs convolutional neural networks enhanced with an attention mechanism. TCDF identifies causal relationships by learning which time series and their corresponding lags contribute most to predicting a target series, effectively capturing complex temporal patterns while providing interpretable lag-specific causal inference.

Recently, Kolmogorov-Arnold Networks (KANs) were introduced by Liu et al. \cite{liu2024kan}, inspired by the Kolmogorov-Arnold representation theorem \cite{kolmogorov1961representation, braun2009constructive}. KANs have gained significant attention for their accuracy and interpretability. Unlike traditional MLPs, where connections between neurons are linear weights, KANs replace these connections with learnable univariate spline functions, allowing for greater flexibility in activation functions. This design enables KANs to model complex, nonlinear patterns effectively. One of the key features of KANs is their ability to achieve sparsification. The authors introduced a regularisation combining L1 norm and entropy norm on all the splines. As a result, connections with weak contribution can effectively be pruned, i.e. removed, during training. 

Several studies have highlighted the potential of KANs in time series forecasting. Xu et al. \cite{xu2024kolmogorov} developed T-KAN and MT-KAN for time series prediction. They demonstrated that with dynamic activation functions, KANs are particularly effective at capturing complex temporal patterns in time series. Aca-Rubio et al. \cite{vaca2024kolmogorov} showed that KAN not only has better forecasting performance on satellite traffic data than MLP, it also shows higher parameter efficiency. 

In this work, we introduce Granger Causality KAN (GC-KAN), a framework for multivariate Granger causality detection that is both accurate and interpretable. Our method builds upon the concept of the cMLP, however, instead of using MLPs for each target time series, we replace them with KANs. To achieve automatic identification of Granger non-causal inputs, we implement an additional proximal operator during training, applied only to the edges connecting the inputs and the first hidden layer. This ensures that irrelevant input features are assigned exact zero weights, allowing Granger causality to be inferred directly from inputs with non-zero weights. The flexibility of KANs' activation functions, combined with the sparsity-inducing regularisation, enhances the model's interpretability. 

In Section \ref{method} we present the mathematical foundation of multivariate Granger causality and detail its implementation in both cMLP and GC-KAN. This is followed by a comprehensive description of the GC-KAN framwork and its training tailored for Granger causality detection. In Section \ref{re} we showcase the experimental results, where we compare the model performance of GC-KAN against cMLP with a specific structured penalty on linear VAR data and nonlinear Lorenz-96 datasets.

\section{Methodology} \label{method}

\subsection{Multivariate Granger Causality}
Multivariate Granger causality determines if one or more time series can predict another within a multivariate framework. In its linear form, it  typically uses the VAR model, assuming linear interactions and stationarity. For an $n-dimensional$ multivariate time series $\mathbf{x}_t \in \mathbb{R}^n$ with $T$ time steps, \(t=1.\ldots,T\), each component in the VAR model of order $p$ is formulated as: 
\begin{equation}\label{linear_VAR}
    {x}_{t,i} = \sum^p_{k=1} A^{(k)}_i\mathbf{x}_{t-k} + \mathbf{e}_t,
\end{equation}
where $A_{i}^{(k)}$ are ($1\times n$) vectors representing linear coefficients for each lag $k$ and $\mathbf{e}_t$ is the vector of noise with zero mean and constant variance. Granger causality from $x_{t,j}$ to $x_{t,i}$ is inferred by testing the null hypothesis 
\begin{equation}
    H_0 : (A_{i}^{(k)})_j = 0 \quad \text{for}\; j\not=i, \;\;\text{and all} \quad k = 1, \dots, p.
\end{equation}

Nonlinear Granger causality relaxes the linearity assumption, allowing for more complex dependencies. The VAR model in Eq. \ref{linear_VAR} is generalised by replacing the summation with a nonlinear function \(f_i(\cdot)\), which reads
\begin{equation}\label{nonlinear_VAR}
    {x}_{t,i} = f_i(A_i^{(1)}\mathbf{x}_{t-1},A_i^{(2)}\mathbf{x}_{t-2}, \dots,A_i^{(p)}\mathbf{x}_{t-p}) + \mathbf{e}_t.
\end{equation}
where, in this case, \(A_i^{(k)}\) are \(n\times n\) diagonal matrices. 
In practice, such function is usually estimated via neural networks.

In the nonlinear case, Granger causality is identified if including past values of a particular series in $f$ improves predictive accuracy for the target series. Mathematically, similarly with the previous case, the null hypothesis for testing Granger causality from $x_{t,j}$ to $x_{t,i}$ can be expressed as:
\begin{equation}\label{non granger non linear}
    H_0 : (A_i)_{j,j}^{(k)} = 0 \quad \text{for}\; j\not=i, \;\;\text{and all} \quad k = 1, \dots, p.
\end{equation}
Under this null hypothesis \( H_0 \), the past values of \( x_{t,j} \) do not contribute to the prediction of \( x_{t,i} \), implying no Granger causality from \( x_{t,j} \) to \( x_{t,i} \).

In practice, linear VAR and nonlinear VAR models are intuitive approaches to identify Granger causality between variables within a system. However, several assumptions are required for these models to be an appropriate framework \cite{assaad2022survey}. Firstly, all the observed time series are assumed to be stationary and form a complete system. Stationarity means the statistical properties of all series do not change over time. The assumption of a complete system implies that there are no unobserved confounders \cite{vanderweele2013definition}. These conditions ensure that the detected relationships are consistent and without bias.

Moreover, all time series are assumed to have identical, discrete frequencies that align with true causal lags so that the modelled relationships can accurately reflect the true dependencies \cite{shojaie2022granger}. The underlying lagged dependencies are assumed to have a finite order, such lag order $p$ must be appropriately chosen to capture the dynamics of the system. Selecting an insufficient lag order may lead to omitted causal links, while an overly large lag order increases model complexity and the risk of overfitting. 

Both the linear and nonlinear frameworks offer powerful tools to analyze Granger causality, with linear models providing simplicity and interpretability, while nonlinear models enable the detection of complex functional dependencies.

\subsection{cMLP for Granger Causality}

The cMLP framework introduced in \cite{Tank_2021} builds on the nonlinear Granger causality concept by using neural networks to estimate the function \(f_i(\cdot)\) in Eq.~\ref{nonlinear_VAR}. 
It addresses the challenge of detecting nonlinear causal relationships in high-dimensional time series by combining predictive modeling with structured regularization, enabling the identification of sparse and interpretable causal pattern.

For a system described by Eq.~\ref{nonlinear_VAR}, the cMLP model trains one feedforward neural network for each target time series in \(\mathbf{x}_t\). The inputs of each MLP are the lagged observations of all variables up to a max lag \(p\), forming a vector of size \(n \times p\). Each MLP is tasked with predicting the value of a specific target series \(x_{t,i}\) based on these inputs. Since the lagged input vector requires \(p\) previous time steps, the loss function starts at \(t = p\) to ensure that all lagged inputs are available. 

A critical innovation in the cMLP framework lies in how Granger causality is extracted from the trained networks. The first-layer weights of each MLP, which connect the input variables to the hidden layer, are subjected to sparsity-inducing regularization during training. This regularization ensures that only the most relevant input variables retain nonzero weights, with irrelevant variables being effectively pruned. The presence of non-zero weights after training indicates that the corresponding lagged inputs contribute to the prediction of the target series, hence revealing Granger causal relationships.

Mathematically, the cMLP loss function consists of two main components: the mean squared error (MSE) for prediction accuracy and a sparsity-inducing penalty applied to the first-layer weights to shrink the entire set of first layer weights for input series $j$ to zero. The loss function is given by:
\begin{equation}
    \mathcal{L} = \sum_{t=p}^T \left( x_{t,i} - f_i \left( \mathbf{x}_{(t-1):(t-p)} \right) \right)^2 
    + \lambda \sum_{j=1}^n \Omega(W^{1}_{:j}),
\end{equation}
where \( W^{1}_{:j} \) denotes the weights associated with the \(j\)-th input variable across all lags.

The three penalties that the cMLP framework incorporates are Group Lasso (GL) \cite{yuan2006model}, Hierarchical Group Lasso (H) \cite{nicholson2020high} and the Group Sparse Group Lasso (GSGL). It is demonstrated in \cite{Tank_2021} that the hierarchical penalty outperformed the other two across different max lag values, showing robustness to input lag. In this study, we focus on the H penalty, and the results of GC-KAN are compared against cMLP with H in Section. \ref{re}.

The Hierarchical Group Lasso penalty imposes a structured sparsity constraint across time lags. It performs simultaneous selection of Granger causal variables and their maximum time lag order. Specifically,
it ensures that higher lags of a variable can only contribute to the prediction if all lower lags are also included. 
The penalty is defined as:
\begin{equation}\label{h}
    \Omega_{\text{H}}\left( \mathbf{W}^{(1)} \right) = \sum_{j=1}^n \sum_{k=1}^p \| \mathbf{W}_{j, k:p}^{(1)} \|_F,
\end{equation}
where \(\mathbf{W}_{j,k:p}^{(1)}\) includes all weights connecting lags $k$ through $p$ of variable \(j\) to the hidden layer, $\| \cdot \|_F$ is the Frobenius norm which penalises the magnitude of the grouped weights.

The cMLP model is trained using proximal gradient descent, which alternates between a gradient descent step to minimize the smooth part of the loss function (MSE) and a proximal update to enforce sparsity in the first-layer weights. 
By combining these techniques, the cMLP framework produces sparse and interpretable models that identify the nonlinear causal relationships driving the target time series. This makes it particularly effective for high-dimensional systems where traditional Granger causality methods struggle to scale.

\subsection{GC-KAN for Granger Causality}

Building upon the structure of the cMLP, our GC-KAN framework models each time series individually using a KAN, enabling flexible and interpretable causality detection.

Instead of linear weights in MLPs, KANs incorporate learnable, spline-based activation functions on the edges connecting neurons. Specifically, each connection between neurons is parameterized by a combination of base weight \( W_{\text{base}} \) and spline weight \( W_{\text{spline}} \) constructed using B-splines \cite{de2002spline}.
Given an input vector \( \mathbf{z} \), the output of the first hidden layer is defined as:
\begin{align}
    \mathbf{h}^1 &= \Phi_0 \mathbf{z}, \label{equ:KAN_first_layer} \\
     \Phi_0 &= W_{\text{base}} \sigma(\mathbf{z}) + W_{spline}spline(\mathbf{z}).
\end{align}
where \( \sigma(\cdot) \) is the basis function (e.g., SiLU). The spline is computed using B-spline basis functions, which are learnable during training. The output of deeper layers follows a similar formulation, hence the network architecture with $l$ layers can be described as:
\begin{align}\label{KAN_equation}
    {\rm KAN}(\mathbf{z}) &= f(\mathbf{z}) \\
    &= (\Phi_{l-1} \circ \Phi_{l-2} \circ \cdots \circ \Phi_0)(\mathbf{z}).
\end{align}

These components allow KANs to adaptively capture localized nonlinear interactions while favoring interpretability. Such flexible architecture enables KANs to model both linear and nonlinear dependencies in the data, making them particularly well-suited for Granger causality analysis.

\subsubsection{Mathematical Formulation for GC-KAN}

The GC-KAN model consists of one KAN for each time series in the multivariate system \(\mathbf{x}_t \). Each KAN is tasked with predicting a single target series \(x_{t,i}\) based on the lagged observations of all series. The input vector for each GC-KAN is defined as:
\[
\mathbf{z} = \left[ \mathbf{x}_{t-1}, \mathbf{x}_{t-2}, \dots, \mathbf{x}_{t-p} \right] \in \mathbb{R}^{np \times 1}.
\]
The output of each KAN is the predicted value of the corresponding target series \(x_{t,i}\):
\[
\hat{x}_{t,i} = f_i(\mathbf{z}),
\]
where \(f_i\) is the function learned by the \(i\)-th KAN.

The KANs employ two regularization techniques to promote sparsity and improve interpretability: an L1 norm penalty and an entropy-based penalty. Formally, the L1 norm for activation function $\phi$ is defined as the average magnitude over its $N$ inputs:
\begin{equation}\label{equ:GC-KAN_phi_norm}
    |\phi|_1 = \frac{1}{N}\sum^{N}_{s=1}|\phi(x^{(s)})|,
\end{equation}
and for the KAN layer $\Phi$ with $n_{\rm in}$ inputs and $n_{\rm out}$ outputs, the L1 norm is the sum of Eq. \ref{equ:GC-KAN_phi_norm}
\begin{equation}
    |\Phi|_1 = \sum^{n_{\rm in}}_{i=1} \sum^{n_{\rm out}}_{j=1} |\phi_{i, j}|_1.
\end{equation}
The entropy regularization is the entropy of $\Phi$, which is
\begin{equation}
    S(\Phi) = - \sum^{n_{\rm in}}_{i=1} \sum^{n_{\rm out}}_{j=1} \frac{|\phi_{i, j}|_1}{|\Phi|_1} \log(\frac{|\phi_{i, j}|_1}{|\Phi|_1}). 
\end{equation}
Therefore, the loss function $\mathcal{L}_{\rm GC-KAN}$ of GC-KAN is the sum of MSE loss for prediction and regularisation which induces sparsity, it is given by
\begin{equation}\label{equ:GC-KAN_loss}
    \mathcal{L}_{\rm GC-KAN} =   \mathcal{L}_{\rm MSE}+ \lambda (\mu_1 |\Phi|_1 + \mu_2 S(\Phi)), 
\end{equation}
where $\mu_1$ and $\mu_2$ are the magnitudes to balance the L1 norm and the entropy, and are usually set to $\mu_1 = \mu_2 = 1$. $\lambda$ is the hyper-parameter used to constrain the regularization degree.   

A notable difference between causality extraction in GC-KAN and cMLP with the H penalty is that all lagged inputs in GC-KAN are treated independently, regardless of whether they originated from the same time series. In cMLP with the H penalty, inputs are grouped by variable, with all lags of a given variable forming a hierarchy. This grouping imposes a structured dependency among the lags of the same variable.

\subsubsection{GC-KAN Optimization for Granger Causality}

The first layer of GC-KAN maps time-lagged inputs to hidden representations. Granger causality is extracted from the functional mappings in the first layer, which encode the contribution of each lagged input series to the prediction of the target series. By analyzing these mappings, significant contributions can be identified, representing causal relationships between input features and the target variable.

While KANs trained solely with gradient descent can identify contributions, noise or irrelevant inputs may still influence the results because KAN aims to express the target series in terms of all available inputs. To address this, we propose adding a proximal update step after the gradient descent update.

To achieve this, the GC-KAN training process involves two key steps for optimizing the loss function \(\mathcal{L}_{\text{GC-KAN}}\).

\paragraph{Gradient Descent on the Total Loss }
   The smooth prediction loss (mean squared error) and regularization terms are optimized jointly using gradient descent across all layers:
\begin{equation}
    \Phi \leftarrow \Phi - \eta \nabla \mathcal{L}_{\text{GC-KAN}},
\end{equation}
where \(\eta\) is the learning rate.

This step ensures that the network learns both predictive accuracy and sparsity-driven representations simultaneously.

\paragraph{Proximal Operator for the First Layer}
   To enforce sparsity in the functional mappings of the first layer, a proximal operator is applied to the weights \( w_{base} \) and \( w_{spline} \) after gradient descent:
\begin{equation}
    w_{base} \leftarrow \text{Prox}_{\lambda_{prox}}(w_{base}), \quad w_{spline} \leftarrow \text{Prox}_{\lambda_{prox}}(w_{spline}),
\end{equation}
where the proximal operator is defined as:
\begin{equation}
    \text{Prox}_{\lambda_{prox}}(w) = \text{sign}(w) \cdot \max(|w| - \eta \lambda_{prox}, 0).
\end{equation}

This soft-thresholding operation reduces the magnitude of both weights by \( \eta \lambda_{prox} \), setting values below the threshold to zero and preserving only significant contributions. Importantly, this step is applied exclusively to the first layer, as it directly encodes the Granger causal relationships. Penalizing both \( w_{base} \) and \( w_{spline} \) ensures that the total contribution of each lagged input is explicitly regularized, leading to a more interpretable representation of causal structure.
 After training, the total contribution of each lag $k$ for variable $j$ is computed by summing the corresponding weights:
\begin{equation}
    C_{j,k} = \sum^{n_{\rm hidden}}_{i=1} |\phi_{j, k,i}|,
\end{equation}
where $\phi_{j, k,i}$ represents the mapping from lag $k$ of input series $j$ to the $i$-th neuron in the hidden layer. Fig. \ref{prox}, in the Appendix, illustrates the difference between the standard training and training with proximal operator.  


The intuition behind this proximal step is to address a limitation of the original training and pruning approach. While the standard method can help identify Granger causal signals in the inputs, the weights associated with irrelevant inputs are not set exactly to zero, hence a threshold is required manually to prune out unwanted features. Such training does provide insights into contributions of features at different lags, but it does not automatically reveal Granger causality. By combining gradient descent for the entire network with a proximal operator for the first layer, GC-KAN balances predictive performance and causal interpretability, allowing for clear identification of Granger causal relationships.

\section{Results}\label{re}

We compare the performance between cMLP and GC-KAN with synthetic dataset generated with VAR models and Lorenz-96 with known causal relations. The comparison is performed in two aspects, causal detection accuracy using area under ROC (AUROC) and qualitative analysis of lag selection. For both experiments, the cMLP for comparison has 1 hidden layer with 100 hidden neurons, and the penalty is Hierarchical Group Lasso. It is showed in \cite{Tank_2021} that such model setup produces the best AUROC score for cMLP in both VAR and Lorenz-96 cases.

\subsection{VAR model}
\subsubsection{Granger causality detection}

For linear Granger causality where the underlying dynamics can be represented using a VAR model, we simulate data from $n=20$ VAR(1) and VAR(2) models. Each time series generated has self dependencies and three randomly selected parents among the other $n-1$ series. The influence of a parent series on a target series is represented by uniform coefficients at lag 1 for VAR(1) and at lag 1 and 2 for VAR(2) models, while the coefficients of all other lagged relationships are set to zero.

In experiments with VAR models, the following GC-KAN structure was implemented: $n\times p$ input features representing lagged time series, 1 hidden layer with 1 hidden neuron, 1 output neuron representing the target series. The choice of 1 hidden neuron was made for simplicity and interpretability. Our experiments showed that using more hidden neurons did not significantly improve performance, as the relationships in the VAR data are linear and sparse. For VAR data, the target series is a linear combination of a few lagged inputs, which, in theory, can be represented using sum of linear weights. Thus, additional hidden neurons may not be necessary, as they introduce redundancy for such linear relations. Besides observations in our experiments, a few examples in \cite{liu2024kan} also demonstrate the sufficiency of a single hidden layer with one hidden neuron for similar tasks. Both cMLP and GC-KAN are initialized with max lag $p = 5$. 

We compare the results of GC-KAN and cMLP across three model setups with varying time series lengths. Model evaluation is conducted using the average AUROC over 5 runs. To evaluate Granger causality, models are trained with a range of proximal strengths, and the resulting binary Granger causality matrix for each model is compared against the ground truth matrix to compute the true positive rate (TPR) and false positive rate (FPR). The list of sensitivities is then plotted on an ROC curve. The results are shown in Table \ref{tab:GC_GC-KAN_cMLP} for three different time series length $T \in (250,500,1000)$. The performance of the GC-KAN and cMLP models is quite similar when $T \geq 500$. For both VAR(1) and VAR(2) with 1000 sample size, GC-KAN outperformed cMLP by a slight margin. However, it is observed that GC-KAN struggles more than cMLP with lower sample size. Both show better detection accuracy with increase in sample size.

\begin{table}[h]
    \centering
    \renewcommand{\arraystretch}{1.3}
    \scalebox{1.1}{
    \begin{tabular}{|c|c|c|c|}
        \hline
        \multicolumn{4}{|c|}{\textit{VAR(1)}} \\ 
        \hline
        Model & T=250 & T=500 & T=1000 \\ 
        \hline
        cMLP & $91.6\pm0.4$ & $94.9\pm0.2$ & $98.4\pm0.1$ \\ 
        
        \textbf{GC-KAN} & $81.7\pm0.2$ & $96.5\pm0.1$ & $99.3\pm0.1$ \\ 
        \hline
        \multicolumn{4}{|c|}{\textit{VAR(2)}} \\ 
        \hline
        Model & T=250 & T=500 & T=1000 \\ 
        \hline
        cMLP & $84.4\pm0.2$ & $88.3\pm0.4$ & $95.1\pm0.2$ \\ 
        
        \textbf{GC-KAN} & $76.1\pm0.3$ & $84.5\pm0.3$ & $95.9\pm0.1$ \\ 
        \hline
        
    \end{tabular}}
    \caption{Comparison of AUROC for Granger causality selection between GC-KAN and cMLP. Three VAR models with order 1 and 2 have been implemented. The average performance of 5 initialisations for different time series length are shown. The results for cMLP are taken directly from \cite{Tank_2021}.}
    \label{tab:GC_GC-KAN_cMLP}
\end{table}

\subsubsection{Lag selection comparison}

\begin{figure}[t]
    \centering

    \begin{subfigure}[t]{0.35\textwidth}
        \centering
        \includegraphics[width=\textwidth]{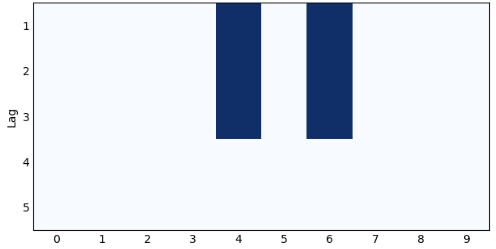}
        \caption{Ground truth for Granger causality.}
        \label{fig:groundtruth}
    \end{subfigure}
    
    \begin{subfigure}[t]{0.35\textwidth}
        \centering
        \includegraphics[width=\textwidth]{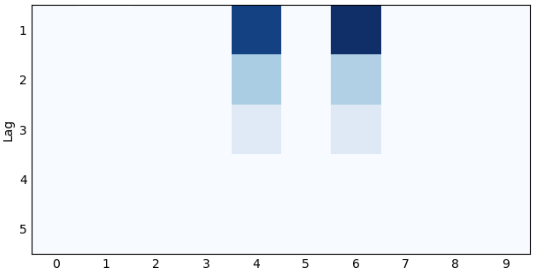}
        \caption{cMLP with Hierarchical Group Lasso penalty.}
        \label{fig:cmlp_detection}
    \end{subfigure}

    \begin{subfigure}[t]{0.35\textwidth}
        \centering
        \includegraphics[width=\textwidth]{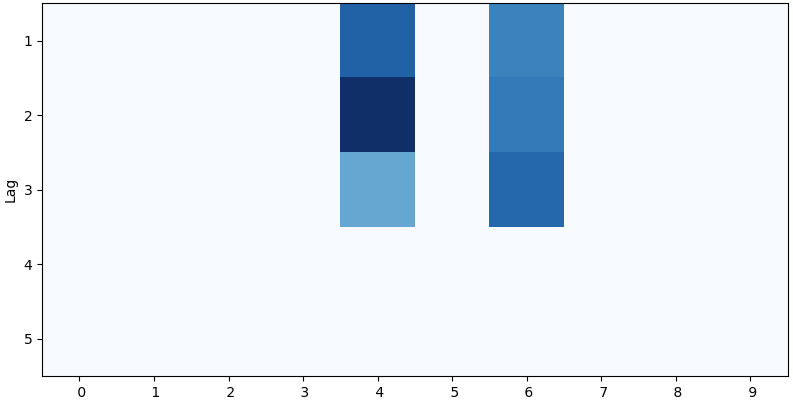}
        \caption{GC-KAN.}
        \label{fig:GC-KAN_detection}
    \end{subfigure}
    
    \vspace{0.5em}
    \caption{Example comparison of Granger causality detection by cMLP with H penalty and GC-KAN against the ground truth. Data is generated from $n = 10$ VAR(3) model. The plots illustrate results from a single cMLP and GC-KAN, focusing on a specific target series. The x-axis represents the input series, while the y-axis corresponds to the input lags. Plots (b) and (c) demonstrate that both cMLP and GC-KAN correctly identify the Granger causal parents of the target series, as indicated by the ground truth in (a). However, the two models differ in their distribution of contributions across different lags, with GC-KAN showing a more even contribution profile compared to cMLP.}
    \label{fig:cmlp_vs_groundtruth}
\end{figure}

For cMLP with the Hierarchical Group Lasso penalty, as expressed in Eq.\ref{h}, the regularization simultaneously selects Granger causal variables and determines the maximum lag order of the interaction. This penalty ensures that for each variable, there exists a maximum lag $\tilde{k}$ such that all weights associated with lags greater than $\tilde{k}$ are zero, while weights for all lags less than or equal to $\tilde{k}$ are non-zero if the variable is relevant. This structure enforces a hierarchy among lags, allowing higher lags to contribute only if all lower lags are already active. During the optimization process, lower lags are prioritized to minimize the penalty. Fig. \ref{fig:cmlp_detection} illustrates an example of the contribution detected by the Hierarchical Group Lasso for each lag. Clearly, $k=1$ has much higher contribution than $k=3$ for both parents, it shows a smooth decrease in contribution for higher lags. 

On the other hand, with GC-KAN model, the contributions across different lags are more even thanks to the combination of L1 norm and entropy norm in the loss function. The former encourages sparsity at the element level unlike group shrinkage, the entropy regularisation spreads contributions across the spline weights. We observe a more evenly distributed contribution.

\subsection{Lorenz-96}

For nonlinear Granger causality detection, we apply our model to the simulated Lorenz-96 data \cite{karimi2010extensive}, whose relationships are defined by:
\begin{equation}
    \frac{dx_{t,i}}{dt} = \left( x_{t,i+1} - x_{t,i-2} \right) x_{t,i-1} - x_{t,i} + F,
\end{equation}
where F is a forcing constant that determines the level of nonlinearity. The rate of change of $x_{t,i}$, depends nonlinearly on the difference between its next and second previous neighbors, scaled by its previous neighbor, with an additional linear term and a forcing constant. This equation explicitly models the temporal dependencies of a variable $x_{t,i}$ on its neighboring variables over time. In this case, $x_{t,i+1}$, $x_{t,i-2}$ and $x_{t,i-1}$ Granger-cause $x_{t,i}$ as their values influence its future dynamic.

Following the  experimental settings in \cite{Tank_2021}, we simulate Lorenz-96 model with $n=20$ and a sampling rate of 0.05. We evaluate GC-KAN performance based on the AUROC score across three time series lengths $T \in (250,500,1000)$ and two forcing constants $F \in (10,40)$. The GC-KAN model contains 20 identical KANs, each receiving 100 input neurons corresponding the lagged values of all variables ($20 \times \text{max lag} (p = 5) $). Each network contains 1 hidden layer with 10 add hidden neurons and a single output neuron. As demonstrated in \cite{liu2024kan}, KANs effectivily approximate multiplication interactions using the formula $2xy = (x+y)^{2} - (x^{2} + y^{2})$. This motivated our choice of 10 add hidden neuron to capture nonlinear dependencies.

\begin{table}[h]
    \centering
    \renewcommand{\arraystretch}{1.3}
    \scalebox{1.1}{
    \begin{tabular}{|c|c|c|c|}
        \hline
        \multicolumn{4}{|c|}{\textit{F = 10}} \\ 
        \hline
        Model & T=250 & T=500 & T=1000 \\ 
        \hline
        cMLP & $86.6\pm0.2$ & $96.6\pm0.2$ & $98.4\pm0.1$ \\ 
        
        \textbf{GC-KAN} & $86.9\pm0.5$ & $92.1\pm0.3$ & $96.2\pm0.2$ \\ 
        \hline
        \multicolumn{4}{|c|}{\textit{F = 40}} \\ 
        \hline
        Model & T=250 & T=500 & T=1000 \\ 
        \hline
        cMLP & $84\pm0.5$ & $89.6\pm0.2$ & $95.5\pm0.3$ \\ 
        
        \textbf{GC-KAN} & $86.3\pm0.2$ & $87.1\pm0.4$ & $95.7\pm0.2$ \\ 
        \hline
        
    \end{tabular}
    }
    \caption{Lorenz-96 with two different forcing constants and 3 time lengths. Results show the average performance of 5 initialisations for different time series length and forcing constants. The results for cMLP are taken directly from \cite{Tank_2021}. }
    \label{lorenz96}
\end{table}

The results in Table \ref{lorenz96} show a very similar performance between cMLP and GC-KAN across different time series length. Both models show an increasing score as the sample size increases. Surprisingly, in this case, GC-KAN outperformed cMLP in low sample size situation, suggesting spline functions adapt more effectively to nonlinear dependencies with limited data. As an additional evaluation, although the GC-KAN and cMLP leverage two different classes of neural networks, we compare their performance when the same number of hidden neurons are used. According to \cite{Tank_2021}, with 10 hidden neurons, the cMLP with hierarchical Group Lasso achieves an AUROC score of 94\% for a sample size of 1000 and $F=40$, which is approximately 1\% - 2\% lower than GC-KAN. This difference highlights a key advantage of KANs, while both models with 10 hidden neurons have very limited trainable parameters, KANs offer greater flexibility due to their learnable spline functions,  which allows KANs to capture complex nonlinear dynamic more accurately, particularly when the network architecture is small.

\section{Conclusions}
This paper, introduces GC-KAN, a novel framework for Granger causality detection that combines the flexibility and interpretability of Kolmogorov-Arnold Networks with a sparsity-inducing proximal gradient approach. Unlike traditional methods, GC-KAN uses learnable spline-based activation functions to model both linear and nonlinear temporal dependencies, enabling an accurate identification of Granger causal relationships. The proximal operator, applied to the first layer, ensures sparsity, assigning exact zero weights to irrelevant inputs, and facilitating direct detection of Granger causality. We have shown that, by using compact architectures, even down to one hidden neurons only, GC-KAN achieves strong performance, demonstrating that simple network structures are sufficient for uncovering complex Granger causal patterns.

We demonstrated the effectiveness of GC-KAN through extensive experiments on simulated data, including linear VAR models and nonlinear Lorenz-96 systems. For VAR models, GC-KAN and cMLP achieved comparable AUROC scores, with GC-KAN slightly outperforming cMLP in high sample size scenarios but they are slightly less performing with low sample size. In contrast, for the Lorenz-96 system, GC-KAN showed stronger performances in low-sample-size settings, highlighting its adaptability to nonlinear dependencies even with limited data.

In this paper, we discussed a minimal application of KAN to Granger causality detection, demonstrating its potential to outperform existing deep learning methodologies. The GC-KAN framework proposed here opens up several exciting avenues for future research. First, incorporating structured penalties, such as hierarchical group lasso, could enhance GC-KAN’s performance. Additionally, the pruning capabilities of L1 and entropy regularizations could be explored to optimize the KAN architecture for specific datasets, improving both interpretability and training efficiency. Finally, by exploiting KANs’ ability to fit symbolic functions, GC-KAN could be used to derive explicit functional expressions for Granger causal relationships. These symbolic representations could offer deeper insights into the dynamics of complex systems, making GC-KAN a powerful tool for analyzing real-world physical systems.

\newpage

\bibliographystyle{plain}
\bibliography{causality}

\onecolumngrid

\newpage
\section{Appendix}\label{sec:KAN}

\subsection{Training results with and without proximal operator }

\renewcommand{\thefigure}{A.\arabic{figure}} 
\setcounter{figure}{0}
\begin{figure}[h]
    \centering
    \begin{subfigure}[t]{0.45\textwidth}
        \centering
        \includegraphics[width=\textwidth]{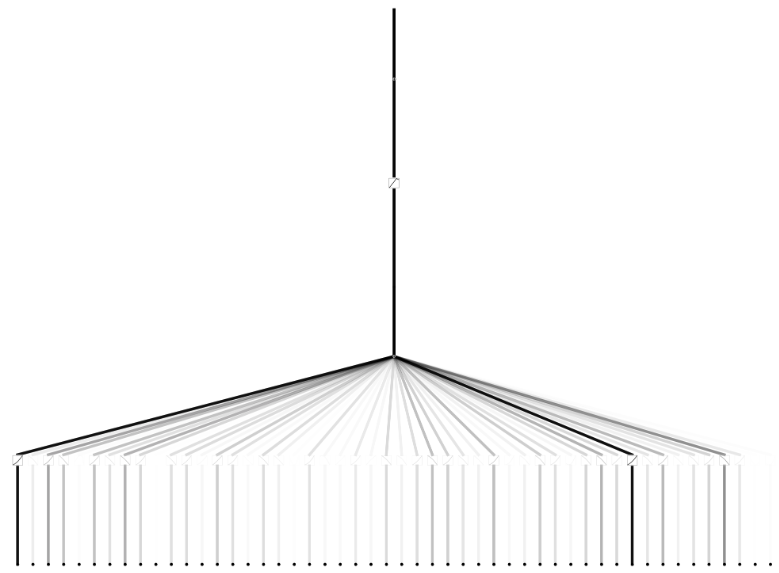}
        \caption{Standard KAN training with LBFGS.}
    \end{subfigure}
    \hfill
    \begin{subfigure}[t]{0.45\textwidth}
        \centering
        \includegraphics[width=\textwidth]{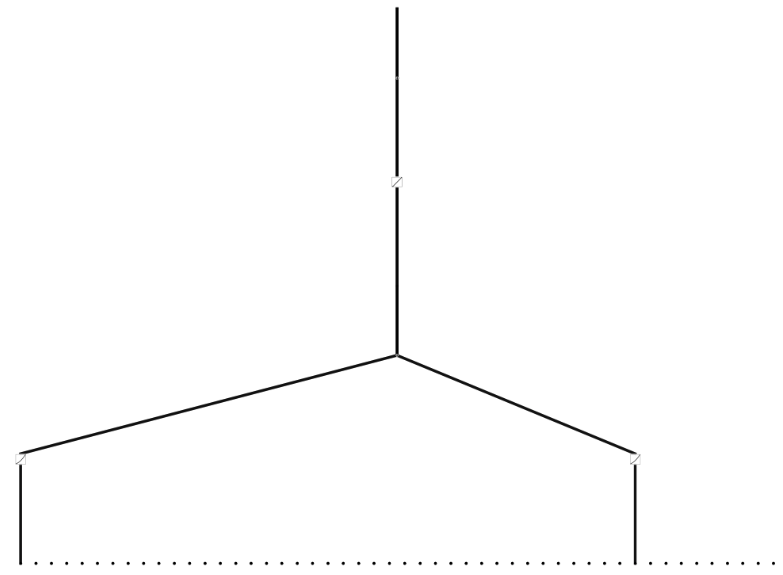}
        \caption{Standard training plus proximal operator applied to the input layer.}
    \label{pro}
    \end{subfigure}
    \caption{Two identical KANs were trained using the same sets of training parameters on the same VAR(1) data with 10 variables and input max lag $p=5$ for a specific target series. Network structure from bottom to top of the plots are [50,1,1] representing 50 inputs, 1 hidden neuron and 1 output neuron. Fig.(a) shows the standard training output, Fig.(b) illustrates the result when KAN is trained with proximal operator applied to the input layers. The darkness of the connecting edges represents the strength of the contribution of the inputs to the output, with darker edges indicating stronger contributions.}
    \label{prox}
\end{figure}

Figure \ref{prox} highlights the differences in training methods. While the KAN trained with the standard LBFGS method successfully identifies significant contributions from the two Granger causal parents, it also assigns non-zero weights to irrelevant inputs. This requries manual thresholding to prune these irrelevant inputs or risks introducing noise that could compromise the accuracy of Granger causality detection. In contrast, as shown in Figure \ref{pro}, the KAN trained with the proximal operator produces a much cleaner structure, where only the Granger causal parents retain non-zero weights. This output can be directly utilized for Granger causality detection without additional post-processing.

\end{document}